\documentclass{elsarticle}
\graphicspath{ {./figures/} }
\usepackage{hyperref}
\usepackage{float}
\usepackage{verbatim}
\usepackage{apalike}
\usepackage{amssymb}
\usepackage{graphicx}
\restylefloat{figure}
\floatstyle{plaintop}
\restylefloat{table}
\usepackage{algorithm}
\usepackage{algpseudocode}
\usepackage{amsmath}

\journal{Expert Systems with Applications}
 
\begin{document}
\begin{frontmatter}
 
 
\title{ShopEase: A Generative AI-Based Multi-Agent Framework for Intelligent Enterprise Customer Support Using Hybrid Retrieval-Augmented Generation}
 
\author[label1]{Aakash Kumar Tiwari \corref{cor1}}
\ead{tiwariaakash1025@kgpian.iitkgp.ac.in}
 
\author[label1]{Somesh Kumar}
\ead{smsh@maths.iitkgp.ac.in}
 
\cortext[cor1]{Corresponding author.}
\address[label1]{Department of Mathematics, Indian Institute of Technology Kharagpur, Kharagpur, West Bengal, India}

 
\begin{abstract}
Enterprise customer support systems must answer customer questions correctly, retrieve the right policy information, use customer context, and pass difficult cases to human agents when needed. This paper presents ShopEase, a Generative AI-based multi-agent framework for enterprise customer support. The system combines six components: Intent, CRM, Memory, Hybrid RAG, Escalation, and Supervisor, and uses LLaMA 3.2 running locally through Ollama for response generation. The retrieval module combines FAISS (dense retrieval) and BM25 (sparse retrieval), and six configurations are evaluated: BM25-only, FAISS-only, Fair RRF, Weighted RRF, RRF with Cross-Encoder, and Top-10 Hybrid with Cross-Encoder. Instead of using a fixed mapping between intent and policy, the policy category is decided directly from the retrieved documents. The system was evaluated on 2632 held-out customer queries across six categories: Refund, Return, Shipping, Cancellation, Damaged Product, and Unknown. FAISS-only achieved the highest accuracy of 85.37\% (2247 correct predictions), closely followed by Weighted RRF at 85.07\%. BM25-only achieved only 55.74\% accuracy. Adding cross-encoder reranking did not improve results: RRF with Cross-Encoder reached 83.24\%, and Top-10 Hybrid with Cross-Encoder reached 81.88\%, while also increasing response latency. Category-level analysis shows strong performance on Shipping, Cancellation, and Return, while Unknown queries remain the main source of errors. Statistical testing using McNemar's test shows no significant difference between FAISS-only and Weighted RRF, though both perform significantly better than Fair RRF and the cross-encoder configurations. Overall, dense retrieval gives the best accuracy on this dataset, and additional reranking adds processing time without improving classification performance.
\end{abstract}

 
\begin{keyword}
Enterprise Customer Support \sep Generative AI \sep Multi-Agent Systems \sep Retrieval-Augmented Generation \sep Hybrid Retrieval \sep Large Language Models \sep Human-in-the-Loop
\end{keyword}

 
\end{frontmatter}

 
\section{Introduction}
Enterprise customer support systems need to answer customer questions correctly and use relevant customer information during the interaction. Traditional chatbot systems can handle common questions, but they may have difficulty when a query requires policy information, customer history, previous conversation context, or human support. A recent review of customer-support chatbots also shows that chatbot systems are widely studied for improving customer service and satisfaction \cite{rahman2024chatbotreview}. However, a complete enterprise support system needs more than response generation.
Retrieval-Augmented Generation (RAG) is widely used to provide external information to language models during response generation \cite{lewis2020rag}. Dense retrieval methods such as DPR use vector representations to find semantically related documents \cite{karpukhin2020dense}, while BM25 uses lexical matching between the query and documents \cite{robertson2009bm25}. These methods have different strengths. Dense retrieval can handle different wording, while lexical retrieval can work well when important terms directly match the policy text. RAG research has also explored methods such as Self-RAG and Corrective RAG to improve the quality of retrieved information \cite{asai2024selfrag, yan2024crag}. However, these approaches mainly focus on the retrieval and generation process rather than the complete enterprise customer-support workflow.
Multi-agent systems provide another way to divide a complex task into smaller components. Recent work has studied the use of multiple agents for query resolution and other AI tasks \cite{wang2025multiagent}. Frameworks such as AutoGen have also shown how multiple language-model agents can work together to solve tasks \cite{wu2024autogen}. However, a customer-support system may also need access to customer records, previous conversations, policy documents, and human intervention. These requirements are not always handled together in a single workflow.
Human involvement is also important for customer-support systems when a query cannot be safely or correctly handled automatically. Human-in-the-loop AI allows human decisions to be included in an AI system \cite{zanzotto2019hitl}. Similarly,  agent-based systems have been studied for handling complex tasks through cooperation between multiple agents \cite{julian2019mas, he2025llmmas}. These studies motivate the use of multiple specialized components instead of relying on a single language model.

\subsection{Research Gap}
Existing studies generally focus on one or two parts of the customer-support problem, such as chatbot response generation, RAG, multi-agent systems, or human-in-the-loop processing. There is a need for a system that combines these components with customer information and conversation memory while also evaluating different retrieval strategies under the same experimental setting.
Another important issue is policy classification. A customer query may use words that are different from the wording in the corresponding policy document.At the same time, some policies may contain similar terms. Therefore, relying only on lexical matching or a fixed mapping between intent and policy may produce incorrect results. A comparison of lexical, dense, hybrid, and reranking-based retrieval methods can provide a clearer view of their performance for enterprise policy retrieval.

\subsection{Motivation}
The main motivation of this work is to build a customer-support system that can use different sources of information before generating a response. Customer information can be obtained from a CRM database, previous messages can provide conversation context, and policy documents can provide the required enterprise information.These inputs can then be used by a Generative AI model to produce the final response.
Based on this motivation, we developed ShopEase, a multi-agent customer-support framework that combines customer context, conversation memory, policy retrieval, response generation, and human escalation. The system uses FAISS and BM25 for retrieval and evaluates different combinations of these methods.The study also examines whether additional RRF and Cross-Encoder stages improve the final policy classification accuracy.

\subsection{Research Contribution}
The main contributions of this work are:

\begin{itemize}
    \item We develop ShopEase, a Generative AI-based multi-agent framework for enterprise customer support that combines customer context, conversation memory, policy retrieval, response generation, and human escalation.
    \item We implement and compare six retrieval configurations: BM25-only, FAISS-only, Fair RRF, Weighted RRF, RRF with Cross-Encoder, and Top-10 Hybrid with Cross-Encoder.
    \item We evaluate the retrieval configurations on a held-out dataset of 2,632 customer queries covering Refund, Return, Shipping, Cancellation, Damaged Product, and Unknown categories.
    \item We analyze the results using accuracy, correct and incorrect predictions, category-level performance, confusion matrix, latency, and statistical significance testing.
    \item We study the effect of removing CRM, Memory, and Escalation components through a component ablation experiment.
\end{itemize}

The rest of the paper describes the related work, proposed methodology, system architecture, experimental setup, results, limitations, and future work.

\section{Literature Review}
Recent work in customer support, retrieval-augmented generation, and multi-agent systems has shown that language models can be used to handle complex user queries. However, these areas have mostly been studied separately. This section reviews the main approaches related to ShopEase.

\subsection{Enterprise Customer Support Systems}
Customer-support chatbots have been widely studied for improving service quality and reducing the workload of support staff.Rahman et al. provide a systematic review of chatbot applications in customer service and discuss their use for customer interaction and satisfaction \cite{rahman2024chatbotreview}. These systems can handle common customer questions, but enterprise support often requires access to customer records, order information, company policies, and previous conversations.
Large language models have improved the ability of chatbots to understand and generate natural language. GPT-3 showed the ability of large language models to perform different language tasks using in-context learning \cite{brown2020gpt3}.  LLaMA 3 further provides open models that can be used for local language-model applications \cite{dubey2024llama3}. However, a language model alone may not have access to current enterprise information. This creates a need for external knowledge retrieval.

\subsection{Retrieval-Augmented Generation}
Retrieval-Augmented Generation combines information retrieval with language generation. Lewis et al. introduced RAG as a method that retrieves external documents and uses them during generation \cite{lewis2020rag}. This approach helps language models use information that is not contained in their model parameters.
Dense retrieval represents queries and documents as vectors and retrieves documents based on semantic similarity. Dense Passage Retrieval (DPR) is an important example of this approach \cite{karpukhin2020dense}. FAISS provides an efficient method for similarity search over dense vectors \cite{johnson2021faiss}.
Lexical retrieval uses the words in the query and documents directly.BM25 is a widely used lexical retrieval method based on term frequency and inverse document frequency \cite{robertson2009bm25}. Information retrieval methods such as these remain useful when important terms in the query directly match the document text \cite{manning2008ir}.
Recent RAG methods have added additional steps to improve retrieval quality. Self-RAG uses retrieval and self-reflection during generation \cite{asai2024selfrag}, while Corrective RAG adds a correction step when the retrieved information is not sufficient \cite{yan2024crag}. Gao et al. provide a broader survey of RAG methods and their main design choices \cite{gao2024ragsurvey}. These works show the importance of retrieval quality, but they do not directly address the complete customer-support workflow used in ShopEase.

\subsection{Multi-Agent Systems}
Multi-agent systems divide a complex task among multiple agents or components.  Earlier work on multi-agent systems studied how different agents can cooperate to solve problems \cite{julian2019mas}. Recent research has extended this idea to large language models. Wang et al. studied multi-agent methods for query resolution, while He et al. reviewed the use of LLMs in multi-agent systems \cite{wang2025multiagent, he2025llmmas}.
Frameworks such as AutoGen provide mechanisms for coordinating multiple language-model agents \cite{wu2024autogen}. ReAct combines reasoning and action to allow language models to interact with external tools \cite{yao2023react}. Toolformer also explored the use of external tools by language models \cite{schick2023toolformer}. These approaches show that separating tasks and using external tools can improve the ability of language-model systems to handle complex tasks.
For enterprise customer support, different tasks can be separated into specialized components. For example, one component can identify the customer request, another can retrieve customer information, and another can retrieve the required policy information. ShopEase follows this idea by using specialized agents for intent, CRM, memory, retrieval, escalation, and workflow control.

\subsection{Human-in-the-Loop Systems}
Fully automatic customer support is not suitable for every situation. Some queries may require human review because of their complexity, uncertainty, or customer-specific requirements. Human-in-the-loop AI includes human decisions as part of the AI workflow \cite{zanzotto2019hitl}.
This idea is also relevant to agentic AI systems,where agents may perform several actions before a final decision is made \cite{murugesan2025agentic, bandi2025rise}. In ShopEase, the Escalation Agent provides a human-in-the-loop path when automatic resolution is not suitable. This allows the system to support both automatic handling and human intervention.

\subsection{Customer Information and Conversation Memory}
Customer support often requires information beyond the current query. Customer profile, order history, customer tier, and previous complaints can affect how a query should be handled. Similarly, previous messages can provide useful context for the current interaction.
Memory is therefore important for maintaining information across a conversation. In a multi-agent system, customer information and conversation history can be provided to the relevant agents before the final response is generated. ShopEase includes separate CRM and Memory components for these two sources of context.

\subsection{Comparison of Existing Approaches}
The reviewed studies address different parts of the customer-support and retrieval problem. Some focus on customer-support chatbots, while others study retrieval, multi-agent systems, or human-in-the-loop AI. Table~\ref{tab:existing_comparison} compares these approaches with ShopEase using the main components relevant to the proposed system.
\begin{table}[H]
\centering
\caption{Comparison of Existing Approaches with ShopEase}
\label{tab:existing_comparison}
\scriptsize
\resizebox{\columnwidth}{!}{%
\begin{tabular}{lcccccc}
\hline
\textbf{Approach} &
\textbf{Support} &
\textbf{Retrieval} &
\textbf{Multi-Agent} &
\textbf{CRM} &
\textbf{Memory} &
\textbf{HITL} \\
\hline
\cite{rahman2024chatbotreview}
& \checkmark & -- & -- & -- & -- & -- \\
\cite{lewis2020rag}
& -- & \checkmark & -- & -- & -- & -- \\
\cite{karpukhin2020dense}
& -- & \checkmark & -- & -- & -- & -- \\
\cite{robertson2009bm25}
& -- & \checkmark & -- & -- & -- & -- \\
\cite{wu2024autogen}
& -- & -- & \checkmark & -- & -- & -- \\
\cite{wang2025multiagent}
& -- & -- & \checkmark & -- & -- & -- \\
\cite{asai2024selfrag}
& -- & \checkmark & -- & -- & -- & -- \\
\cite{zanzotto2019hitl}
& -- & -- & -- & -- & -- & \checkmark \\
\textbf{ShopEase}
& \checkmark & \checkmark & \checkmark &
\checkmark & \checkmark & \checkmark \\
\hline
\end{tabular}%
}
\end{table}
Here, ``--'' indicates that the corresponding capability was not reported or addressed in the cited work. The comparison is based on the capabilities discussed in the cited studies.
The comparison shows that the existing studies mainly address individual parts of the problem. Customer-support studies focus on support interaction, RAG studies focus on external knowledge retrieval, multi-agent studies focus on task coordination, and human-in-the-loop work focuses on human involvement. ShopEase combines these components with customer information and conversation memory in one customer-support workflow.
In addition to combining these components, ShopEase evaluates six retrieval configurations on the same held-out dataset. This provides a direct comparison of BM25, FAISS, RRF, and Cross-Encoder-based retrieval within the same enterprise customer-support setting.
\section{Proposed Methodology}
ShopEase is designed as a multi-agent customer-support system that combines customer information, conversation history, policy retrieval, and human escalation. The workflow takes a customer query as input and processes it through different
components before generating the final response. Each component has a specific role, while the Supervisor coordinates the complete workflow \cite{su2025learnbyinteract, yang2025agentnet}.

\subsection{System Workflow}
The workflow starts when a customer submits a query. A Guardrail component first checks and preprocesses the input. The Intent Agent then identifies the main intent of the query. Customer information and previous conversation details are obtained from the CRM and Memory components. These details are provided as context for policy retrieval.
The Hybrid RAG component retrieves relevant policy documents using dense and sparse retrieval. FAISS is used for semantic retrieval, while BM25 is used for keyword-based retrieval. The retrieved documents are combined using Reciprocal Rank Fusion (RRF). Different retrieval configurations are evaluated in the experiments.
After retrieval, the system determines the relevant policy category from the retrieved documents and prepares the information required for response generation. If the query requires human support, the Escalation Agent handles the human-in-the-loop step. Otherwise, the Supervisor coordinates the response generation and returns the final answer to the customer.
A Reflection component is also used to review the generated response before the workflow is completed.

\subsection{Main Components}
Table~\ref{tab:method_components} summarizes the main components used in ShopEase.
\begin{table}[H]
\centering
\caption{Main Components of ShopEase}
\label{tab:method_components}
\small
\begin{tabular}{p{2.2cm}p{5.0cm}}
\hline
\textbf{Component} & \textbf{Main Function} \\
\hline
Guardrail & Checks and preprocesses the customer input. \\
Intent Agent & Identifies the main intent of the customer query. \\
CRM Agent & Retrieves customer information from the CRM database. \\
Memory Agent & Provides relevant information from previous conversation history. \\
Hybrid RAG Agent & Retrieves relevant policy documents using FAISS and BM25. \\
Escalation Agent & Handles cases that require human support. \\
Supervisor Agent & Coordinates the workflow and controls the final response process. \\
Reflection & Reviews the generated response before completion. \\
\hline
\end{tabular}
\end{table}

\subsection{Customer Context}
ShopEase uses both customer information and conversation history to provide context for query handling. The CRM Agent accesses the SQLite-based CRM database and retrieves available customer information. The Memory Agent provides relevant information from previous interactions.
These two sources help the system use information about the current customer instead of treating every query as an isolated request. The retrieved context is passed to the later stages of the workflow along with the customer query.

\subsection{Hybrid Policy Retrieval}
The policy retrieval stage uses two retrieval methods. FAISS performs dense retrieval using vector embeddings, while BM25 performs sparse retrieval based on term matching. The embedding model used for dense retrieval is \texttt{nomic-embed-text}.

For a query $q$, FAISS returns documents according to their semantic similarity, while BM25 ranks documents according to their lexical relevance. The two ranked lists can then be combined using Reciprocal Rank Fusion (RRF).
The RRF score for a document $d$ is calculated as

\begin{equation}
RRF(d) =
\sum_{m \in M}
\frac{w_m}{k + rank_m(d)}
\end{equation}

After ranking, the retrieved documents are used to identify the most relevant policy category. The category is selected from the policy information contained in the retrieved documents. This retrieval-based approach avoids using a fixed mapping between query intent and policy category.
where $M$ represents the retrieval methods, $w_m$ is the weight assigned to a method, $rank_m(d)$ is the rank of document $d$ for that method, and $k$ is the ranking constant.
For Fair RRF, equal weights are used for FAISS and BM25. Weighted RRF uses different weights for the two retrieval methods. Cross-encoder reranking is evaluated in separate configurations. The main implementation settings are
summarized in Table~\ref{tab:implementation_settings}.

\subsection{Implementation Settings}
Table~\ref{tab:implementation_settings} summarizes the main implementation settings used in the retrieval and generation pipeline.

\begin{table}[H]
\centering
\caption{Implementation Settings of ShopEase}
\label{tab:implementation_settings}
\small
\begin{tabular}{ll}
\hline
\textbf{Component} & \textbf{Setting} \\
\hline
Language Model & LLaMA 3.2 \\
LLM Runtime & Ollama \\
Embedding Model & \texttt{nomic-embed-text} \\
Dense Retrieval & FAISS \\
Sparse Retrieval & BM25 \\
RRF Method & Reciprocal Rank Fusion \\
RRF Constant ($k$) & 60 \\
Fair RRF Weights & FAISS = 1.0, BM25 = 1.0 \\
Cross-Encoder & \texttt{ms-marco-MiniLM-L-6-v2} \\
CRM Database & SQLite \\
Application Interface & Streamlit \\
\hline
\end{tabular}
\end{table}

The same embedding model and policy collection were used across the retrieval experiments. Fair RRF uses equal weights for the FAISS and BM25 rankings, while Weighted RRF uses different weights. Cross-encoder reranking is applied only in the configurations that include the reranking stage.

\subsection{Retrieval Configurations}
The proposed methodology evaluates multiple retrieval settings to study the effect of dense, sparse, hybrid, and reranked retrieval.

\begin{table}[H]
\centering
\caption{Retrieval Configurations Used in ShopEase}
\label{tab:retrieval_configs}
\small
\begin{tabular}{lccc}
\hline
\textbf{Configuration} & \textbf{FAISS} & \textbf{BM25} & \textbf{Cross-Encoder} \\
\hline
BM25-only & -- & \checkmark & -- \\
FAISS-only & \checkmark & -- & -- \\
Fair RRF & \checkmark & \checkmark & -- \\
Weighted RRF & \checkmark & \checkmark & -- \\
RRF + Cross-Encoder & \checkmark & \checkmark & \checkmark \\
Top-10 Hybrid + Cross-Encoder & \checkmark & \checkmark & \checkmark \\
\hline
\end{tabular}
\end{table}

\subsection{Response and Human Escalation}
After policy retrieval, the system uses the retrieved information and available customer context to prepare the response. LLaMA 3.2 is used locally through Ollama for language generation.
When a query cannot be handled reliably by the automated workflow or requires human support, the Escalation Agent transfers the case to the human-review stage. This allows the system to combine automated response generation with human intervention.

\subsection{Workflow Coordination}
The Supervisor Agent controls the overall execution of the workflow. It maintains the shared state between components and ensures that the output of one stage is available to the next stage. This coordination allows intent information, customer context, conversation history, retrieved policies, escalation status, and generated responses to be handled within one workflow.
The complete execution process can be summarized as:
\begin{equation}
\text{Query}
\rightarrow
\text{Guardrail}
\rightarrow
\text{Intent}
\rightarrow
\text{Context}
\rightarrow
\text{Retrieval}
\rightarrow
\text{Escalation}
\rightarrow
\text{Response}
\rightarrow
\text{Reflection}
\end{equation}
This workflow forms the basis for the experiments described in the following sections.
\section{System Architecture}
\begin{figure}[H]
\centering
\includegraphics[width=\columnwidth]{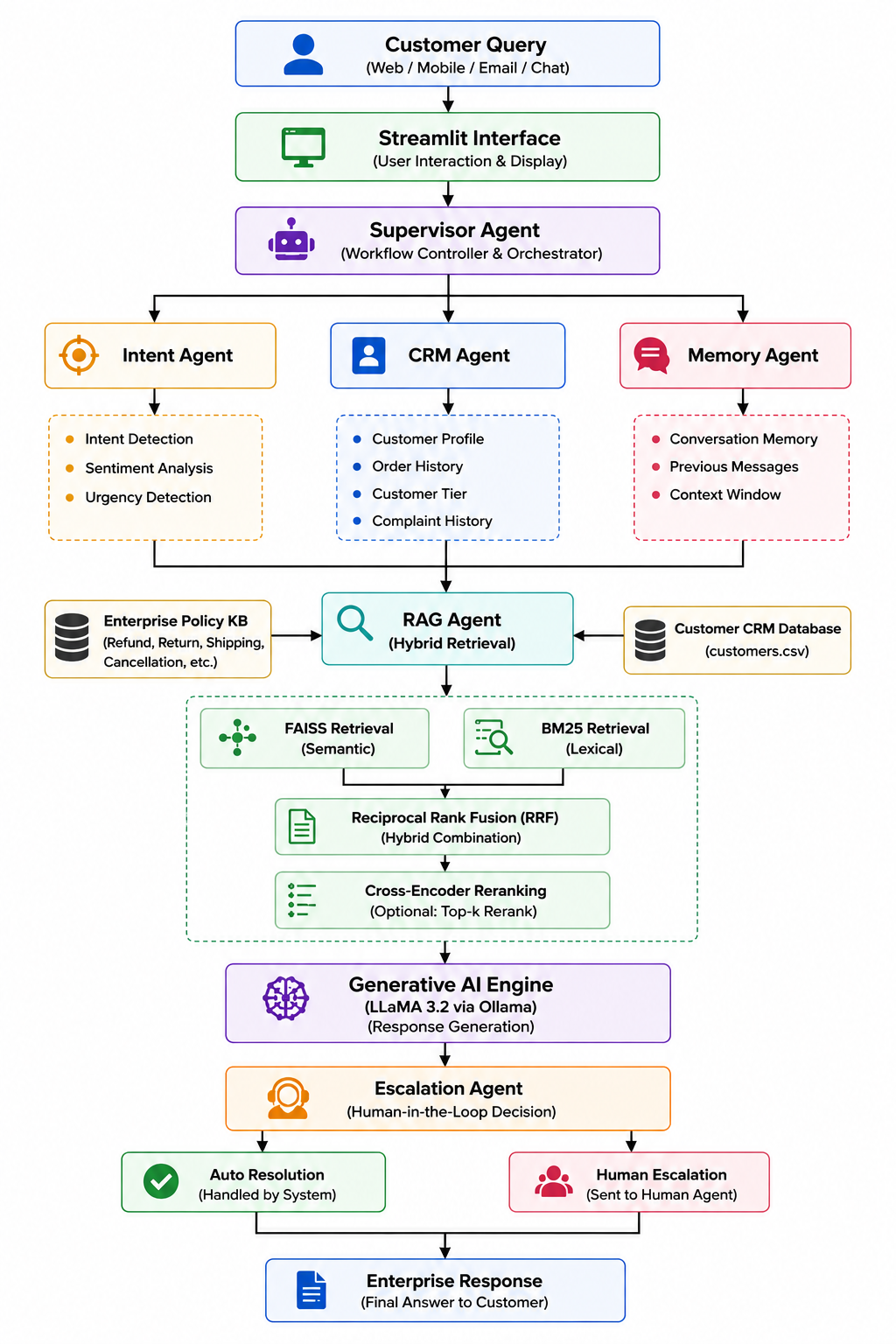}
\caption{Overall architecture of the ShopEase customer-support system.}
\label{fig:architecture}
\end{figure}

The architecture of ShopEase is shown in Fig.~\ref{fig:architecture}. The system is implemented as a graph-based workflow in which different components handle different tasks. The main components include input processing, intent detection,  customer context, policy retrieval, escalation, response generation, and reflection.

\subsection{Input and User Interface}
\begin{figure}[H]
\centering
\includegraphics[width=\columnwidth]{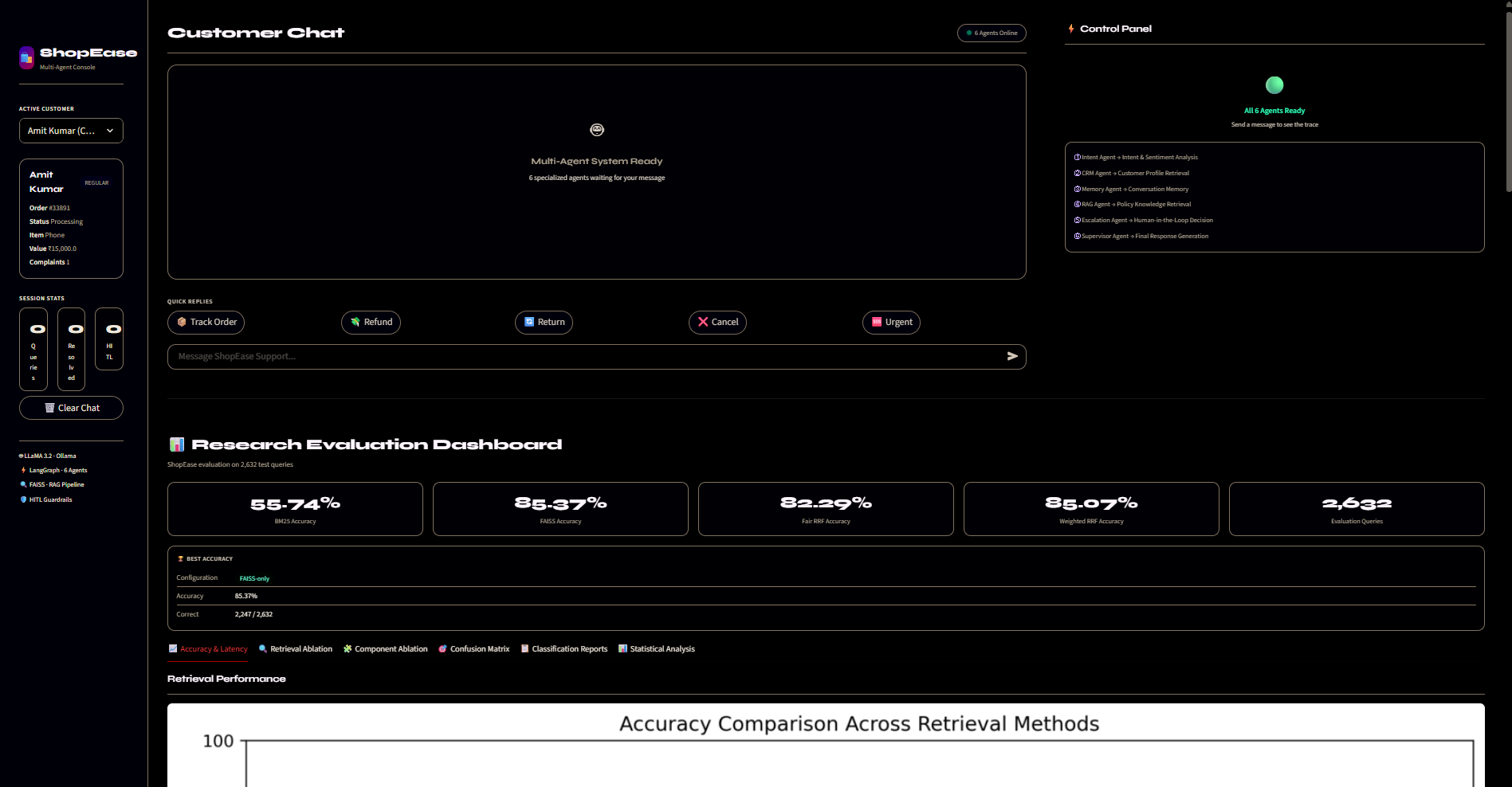}
\caption{Streamlit interface of the ShopEase system.}
\label{fig:streamlit}
\end{figure}

The customer interacts with ShopEase through a Streamlit-based interface. The interface accepts the customer query and displays the generated response along with the relevant workflow information.
Figure~\ref{fig:streamlit} shows the implemented Streamlit dashboard used to interact with the system.

\subsection{Guardrail and Intent Processing}
The Guardrail component processes the incoming query before it enters the main workflow. The Intent Agent then identifies the main intent of the query. The detected intent is stored in the shared workflow state and is available to the later stages.

\subsection{CRM and Memory Components}
The CRM Agent retrieves available customer information from the SQLite-based CRM database. The Memory Agent provides relevant information from previous conversations. These components provide customer and conversation context for the policy retrieval and response generation stages.

\subsection{Hybrid RAG Component}
The Hybrid RAG component retrieves relevant policy information using both dense and sparse retrieval. FAISS is used for dense retrieval with \texttt{nomic-embed-text} embeddings, while BM25 is used for keyword-based retrieval.
The retrieved documents can be combined using Reciprocal Rank Fusion (RRF).The system also supports weighted RRF and cross-encoder reranking. The cross-encoder used in the experiments is \texttt{cross-encoder/ms-marco-MiniLM-L-6-v2}.
The policy category is determined from the retrieved policy documents. Thus, the retrieval stage provides the policy evidence used for the final response rather than relying on a fixed intent-to-policy mapping.

\subsection{Escalation and Supervisor}
The Escalation Agent handles cases that require human support. When escalation is required, the case can be passed to the human-review stage.
The Supervisor Agent coordinates the complete workflow. It manages the shared state and controls the flow of information between the different components. This allows the query, intent, customer context, memory, retrieved policies, and escalation information to be used during response generation.

\subsection{Response Generation and Reflection}
LLaMA 3.2 is used locally through Ollama for response generation. The response is generated using the retrieved policy information and available customer context.
The Reflection component provides a final review step for the generated response before the workflow is completed. The final output is then returned through the Streamlit interface.
\section{Algorithm}
The ShopEase workflow processes a customer query through a sequence of components. The main steps are shown in Algorithm~\ref{alg:shopease}.
\begin{algorithm}
\caption{ShopEase Customer Support Workflow}
\label{alg:shopease}
\small
\begin{tabular}{@{}p{0.35cm}p{7.0cm}@{}}
1. & Receive customer query $q$ \\
2. & Check and preprocess $q$ using Guardrail \\
3. & Identify query intent using Intent Agent \\
4. & Retrieve customer information using CRM Agent \\
5. & Retrieve relevant conversation information using Memory Agent \\
6. & Select the required retrieval configuration \\
7. & Retrieve relevant policy documents using the selected method \\
8. & Determine the relevant policy category from retrieved documents \\
9. & Check whether human escalation is required \\
10. & If escalation is required, use Escalation Agent \\
11. & Otherwise, generate response using policy and customer context \\
12. & Review the generated response using Reflection \\
13. & Supervisor coordinates the final workflow state \\
14. & Return final response $r$ \\
\end{tabular}
\end{algorithm}

\begin{algorithm}[H]
\caption{Hybrid Policy Retrieval with Reciprocal Rank Fusion}
\label{alg:hybrid_retrieval}
\begin{algorithmic}[1]
\Require Query $q$, policy documents $D$
\Ensure Ranked policy documents $D_r$
\State Generate query embedding for $q$
\State Retrieve ranked documents $D_f$ using FAISS
\State Retrieve ranked documents $D_b$ using BM25
\State Initialize RRF score for each document
\For{each document $d$ in $D_f$ and $D_b$}
    \State Compute its RRF score using its rank
\EndFor
\State Combine documents according to their RRF scores
\State Sort documents by decreasing RRF score
\State Apply cross-encoder reranking when enabled
\State Return ranked documents $D_r$
\end{algorithmic}
\end{algorithm}
\section{Experimental Setup}
This section describes the dataset, implementation environment, models, retrieval configurations, and evaluation procedure used to evaluate ShopEase.

\subsection{Evaluation Dataset}
The final evaluation dataset contains 2,632 held-out customer queries. The queries were organized into six policy categories: Refund, Return, Shipping, Cancellation, Damaged Product, and Unknown. The dataset was used only for the final evaluation of the retrieval configurations.
The queries represent common enterprise customer-support cases related to product returns, refunds, shipping, cancellations, and damaged products. The Unknown category contains queries that do not clearly belong to the defined policy categories. This category was included to evaluate how the retrieval system handles queries without a clear policy match.
The same evaluation queries and ground-truth labels were used for all six retrieval configurations. This provides a common evaluation setting and allows a direct comparison of the retrieval methods without changing the test data.
Table~\ref{tab:dataset_distribution} shows the distribution of the evaluation queries.
\begin{table}[H]
\centering
\caption{Distribution of the Evaluation Dataset}
\label{tab:dataset_distribution}
\small
\begin{tabular}{lrr}
\hline
\textbf{Category} & \textbf{Queries} & \textbf{Percentage} \\
\hline
Refund & 481 & 18.27\% \\
Return & 502 & 19.07\% \\
Shipping & 481 & 18.27\% \\
Cancellation & 482 & 18.31\% \\
Damaged Product & 481 & 18.27\% \\
Unknown & 205 & 7.79\% \\
\hline
\textbf{Total} & \textbf{2632} & \textbf{100\%} \\
\hline
\end{tabular}
\end{table}

\subsection{Implementation Environment}
ShopEase was implemented in Python 3.11 on a Windows-based workstation.The application interface was developed using Streamlit. The CRM information is stored in a SQLite database. LLaMA 3.2 is used for response generation through Ollama.  Dense retrieval uses the \texttt{nomic-embed-text} embedding model with FAISS,  while BM25 is used for sparse retrieval.The cross-encoder experiments use
cross-encoder/ms-marco-MiniLM-L-6-v2 \cite{devlin2019bert}.

\subsection{Retrieval Configurations}
Six retrieval configurations were evaluated using the same 2,632 held-out queries. These configurations were selected to compare sparse retrieval, dense retrieval,  hybrid retrieval, and reranking.
\begin{table}[H]
\centering
\caption{Retrieval Configurations Used for Evaluation}
\label{tab:retrieval_configs}
\small
\begin{tabular}{lccc}
\hline
\textbf{Configuration} & \textbf{FAISS} & \textbf{BM25} & \textbf{Cross-Encoder} \\
\hline
BM25-only & -- & \checkmark & -- \\
FAISS-only & \checkmark & -- & -- \\
Fair RRF & \checkmark & \checkmark & -- \\
Weighted RRF & \checkmark & \checkmark & -- \\
RRF + Cross-Encoder & \checkmark & \checkmark & \checkmark \\
Top-10 Hybrid + Cross-Encoder & \checkmark & \checkmark & \checkmark \\
\hline
\end{tabular}
\end{table}
For Fair RRF, FAISS and BM25 are combined with equal weights. Weighted RRF uses different weights for the two retrieval methods. The last two configurations additionally apply cross-encoder reranking.

\subsection{Evaluation Metrics}
The primary evaluation metric is classification accuracy. It is calculated as
\begin{equation}
Accuracy =
\frac{N_{\mathrm{correct}}}{N_{\mathrm{total}}}
\times 100.
\end{equation}
Here, $N_{\mathrm{correct}}$ represents the number of correctly classified queries and $N_{\mathrm{total}}$ represents the total number of evaluation queries.
Precision, recall, F1-score, and support are also used for category-level analysis. Confusion matrices are used to examine the distribution of correct and incorrect predictions across policy categories. Retrieval latency is evaluated using mean, median, minimum, and maximum for the configurations where latency was recorded.

\subsection{Evaluation Procedure}
All six retrieval configurations were evaluated on the same 2,632 held-out queries. For each query, the system retrieved policy information and determined the policy category from the retrieved documents. The predicted category was compared with the ground-truth category.
The number of correct and incorrect predictions was recorded for each configuration. Category-level predictions were also saved for classification reports, confusion matrices, and error analysis.
For statistical comparison, McNemar's test with continuity correction was applied to paired predictions from the same evaluation queries. A significance level of $\alpha=0.05$ was used.
\section{Results and Discussion}
This section presents the experimental results of ShopEase on the 2,632 held-out customer queries. The results are discussed in terms of retrieval accuracy, latency, category-wise performance, error patterns, statistical significance, and component ablation.

\subsection{Overall Retrieval Performance}
\begin{figure}[H]
\centering
\includegraphics[width=\columnwidth]{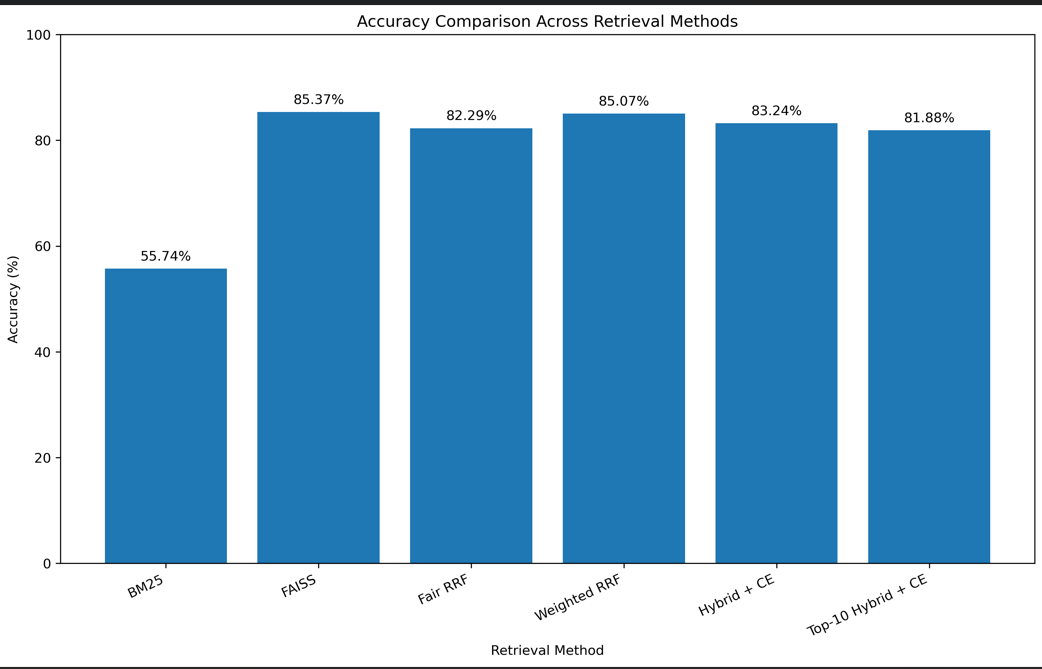}
\caption{Accuracy comparison of the evaluated retrieval configurations.}
\label{fig:accuracy_comparison}
\end{figure}

Table~\ref{tab:overall_results} presents the performance of the six retrieval configurations. FAISS-only gives the highest accuracy of 85.37\%, followed by Weighted RRF with 85.07\%. BM25-only gives the lowest accuracy of 55.74\%.
\begin{table}[H]
\centering
\caption{Overall Retrieval Performance}
\label{tab:overall_results}
\small
\begin{tabular}{lrrr}
\hline
\textbf{Configuration} & \textbf{Correct} & \textbf{Incorrect} & \textbf{Accuracy} \\
\hline
BM25-only & 1467 & 1165 & 55.74\% \\
FAISS-only & 2247 & 385 & 85.37\% \\
Fair RRF & 2166 & 466 & 82.29\% \\
Weighted RRF & 2239 & 393 & 85.07\% \\
RRF + Cross-Encoder & 2191 & 441 & 83.24\% \\
Top-10 Hybrid + Cross-Encoder & 2155 & 477 & 81.88\% \\
\hline
\end{tabular}
\end{table}
Figure~\ref{fig:accuracy_comparison} compares the accuracy of all retrieval configurations.
FAISS-only achieves the highest accuracy of 85.37\%, with 2247 correct predictions out of 2,632 queries. Weighted RRF gives a very close accuracy of 85.07\%, with 2239 correct predictions. The difference between the two configurations is only 0.30
percentage points. BM25-only achieves 55.74\% accuracy, which is 29.63 percentage
points lower than FAISS-only. This indicates that semantic retrieval is more effective than lexical matching for the queries in the evaluated dataset. Fair RRF achieves 82.29\%, which is lower than both FAISS-only and Weighted RRF.
The two cross-encoder configurations also perform below FAISS-only. RRF + Cross-Encoder achieves 83.24\%, while Top-10 Hybrid + Cross-Encoder achieves 81.88\%. Therefore, adding reranking does not improve policy classification accuracy in the
current experiment.

\subsection{Retrieval Latency}
\begin{figure}[H]
\centering
\includegraphics[width=\columnwidth]{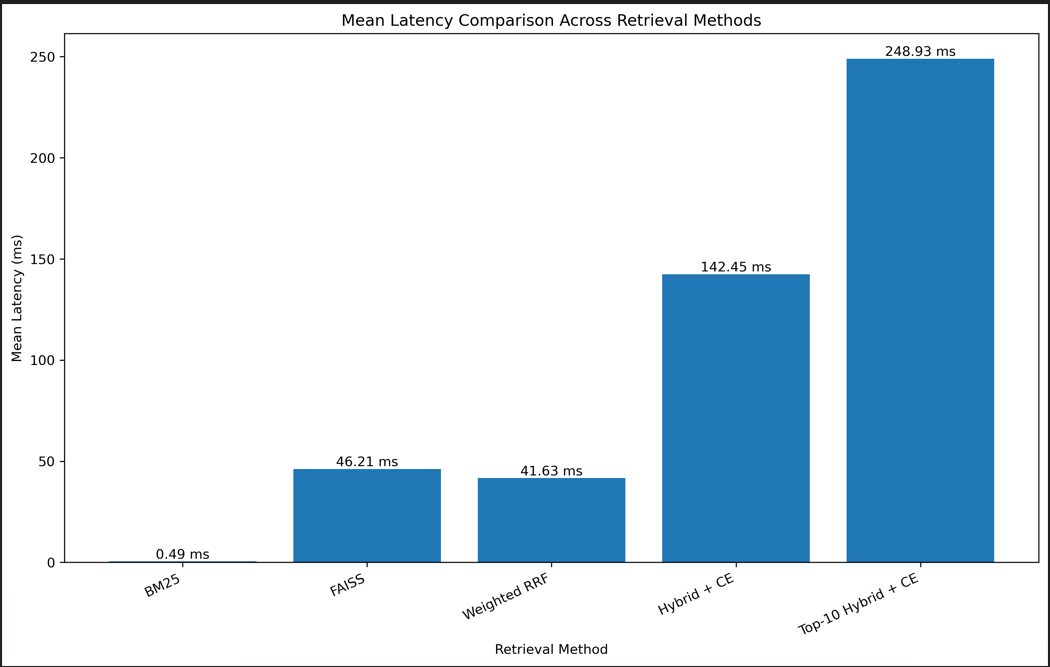}
\caption{Mean retrieval latency of the evaluated configurations.}
\label{fig:latency_comparison}
\end{figure}

Figure~\ref{fig:latency_comparison} shows the mean retrieval latency of the configurations for which latency was recorded.
Table~\ref{tab:latency} gives the detailed latency statistics. BM25-only has the lowest mean latency, while the cross-encoder configurations require more processing time.
\begin{table}[H]
\centering
\caption{Retrieval Latency Statistics}
\label{tab:latency}
\resizebox{\columnwidth}{!}{%
\begin{tabular}{lrrrr}
\hline
\textbf{Configuration} & \textbf{Mean (s)} & \textbf{Median (s)} &
\textbf{Min (s)} & \textbf{Max (s)} \\
\hline
BM25-only & 0.000494 & 0.000410 & 0.000198 & 0.022928 \\
FAISS-only & 0.046205 & 0.044860 & 0.028356 & 0.951277 \\
Fair RRF & Not recorded & -- & -- & -- \\
Weighted RRF & 0.041632 & 0.039754 & 0.025253 & 0.332629 \\
RRF + Cross-Encoder & 0.142449 & 0.111598 & 0.077048 & 15.210859 \\
Top-10 Hybrid + Cross-Encoder & 0.248934 & 0.177453 & 0.092927 & 18.114652 \\
\hline
\end{tabular}%
}
\end{table}
The latency results show a clear cost for the additional reranking stage. BM25-only has the lowest mean latency at 0.000494 seconds, while FAISS-only has a mean latency of 0.046205 seconds. Weighted RRF has a similar mean latency of
0.041632 seconds. Adding the cross-encoder increases the mean latency to
0.142449 seconds for RRF + Cross-Encoder and 0.248934 seconds for Top-10 Hybrid + Cross-Encoder. These configurations also achieve lower accuracy than FAISS-only. Thus, in the current evaluation, the additional reranking time does not result in
better policy classification performance. The latency of Fair RRF was not recorded and is therefore not included in the numerical latency comparison.

\subsection{Category-wise Performance}
FAISS-only is the best-performing configuration and is therefore used for detailed category-wise analysis. Table~\ref{tab:classification_report} presents the category-wise results.
\begin{table}[H]
\centering
\caption{Category-wise Classification Performance of FAISS-only}
\label{tab:classification_report}
\small
\begin{tabular}{lrrrr}
\hline
\textbf{Category} & \textbf{Precision} & \textbf{Recall} & \textbf{F1} & \textbf{Support} \\
\hline
Refund & 95.91\% & 82.95\% & 88.96\% & 481 \\
Return & 82.29\% & 94.42\% & 87.94\% & 502 \\
Shipping & 72.94\% & 99.17\% & 84.05\% & 481 \\
Cancellation & 93.59\% & 96.89\% & 95.21\% & 482 \\
Damaged Product & 89.03\% & 89.40\% & 89.21\% & 481 \\
Unknown & 0.00\% & 0.00\% & 0.00\% & 205 \\
\hline
\textbf{Macro Avg.} & 72.29\% & 77.14\% & 74.23\% & 2632 \\
\textbf{Weighted Avg.} & 79.96\% & 85.37\% & 82.13\% & 2632 \\
\hline
\end{tabular}
\end{table}
The category-wise results show that FAISS performs strongly on most known policy categories. Shipping achieves the highest recall at 99.17\%, followed by Cancellation at 96.89\% and Return at 94.42\%. Cancellation also gives the highest F1-score
among the evaluated categories. Refund has a lower recall of 82.95\%, which is mainly related to confusion with the Return category. The Unknown category remains
the most difficult category, with zero recall and zero F1-score. This shows that the current retrieval system tends to assign unclear queries to one of the known policy categories.

\subsection{Confusion Matrix Analysis}
\begin{figure}[H]
\centering
\includegraphics[width=\columnwidth]{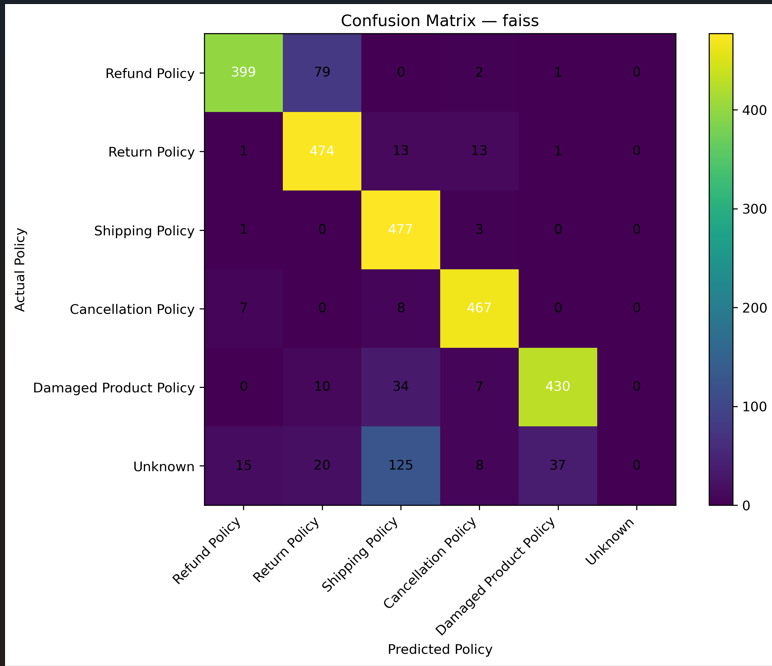}
\caption{Confusion matrix of the FAISS-only configuration.}
\label{fig:faiss_confusion}
\end{figure}

Figure~\ref{fig:faiss_confusion} shows the confusion matrix for the FAISS-only configuration.
The confusion matrix provides a more detailed view of the FAISS results. Shipping has 477 correct predictions out of 481 queries, while Return has 474 correct predictions out of 502 queries. Cancellation also shows strong performance with 467 correct predictions out of 482 queries. The main confusion is between Refund and Return, where 79 Refund queries are classified as Return. Another important pattern is the Unknown category. None of its 205 queries are correctly classified as Unknown. Instead, 125 are classified as Shipping, 37 as Damaged Product, 20 as Return, 15 as Refund, and 8 as Cancellation.
These results suggest that the system handles queries with clear policy-related information well, while queries without a clear policy match remain more difficult.

\subsection{Error Analysis}
Table~\ref{tab:error_analysis} summarizes the main error patterns observed in the FAISS confusion matrix.
\begin{table}[H]
\centering
\caption{Major Error Patterns in FAISS-only Retrieval}
\label{tab:error_analysis}
\small
\begin{tabular}{lrr}
\hline
\textbf{Actual Category} & \textbf{Predicted Category} & \textbf{Errors} \\
\hline
Unknown & Shipping & 125 \\
Unknown & Damaged Product & 37 \\
Unknown & Return & 20 \\
Refund & Return & 79 \\
Damaged Product & Shipping & 34 \\
Return & Shipping & 13 \\
Return & Cancellation & 13 \\
Cancellation & Damaged Product & 13 \\
\hline
\end{tabular}
\end{table}
The error analysis shows two main patterns. First, Refund and Return queries have overlapping terms, which causes some Refund queries to be classified as Return. Second, Unknown queries are often assigned to one of the known policy categories. These errors indicate that queries without a clear policy match remain difficult for the retrieval system.

\subsection{Statistical Significance}
McNemar's test with continuity correction was used to compare the paired
predictions of the retrieval configurations on the same 2,632 evaluation
queries. The significance level was set to $\alpha=0.05$.The results are
shown in Table~\ref{tab:statistical_test}.
\begin{table}[H]
\centering
\caption{Statistical Comparison with FAISS-Only}
\label{tab:statistical_test}
\resizebox{\columnwidth}{!}{%
\begin{tabular}{lccc}
\hline
\textbf{Comparison} & $\boldsymbol{\chi^2}$ & \textbf{$p$-value} & \textbf{Result} \\
\hline
FAISS vs Fair RRF & 19.5719 & 0.000010 & Significant \\
FAISS vs Weighted RRF & 0.6282 & 0.428014 & Not significant \\
FAISS vs RRF + Cross-Encoder & 18.6728 & 0.000016 & Significant \\
\hline
\end{tabular}%
}
\end{table}
The test shows a statistically significant difference between FAISS-only and
Fair RRF ($p=0.000010$). A statistically significant difference is also
observed between FAISS-only and RRF + Cross-Encoder ($p=0.000016$). In
contrast, the difference between FAISS-only and Weighted RRF is not
statistically significant ($p=0.428014$).
The non-significant result for FAISS-only and Weighted RRF is consistent with
their very close accuracy values of 85.37\% and 85.07\%, respectively. The
difference is only 0.30 percentage points. Therefore, although FAISS-only
achieves the highest measured accuracy, its advantage over Weighted RRF is
not statistically significant in this paired evaluation.

\subsection{Component Ablation}
\begin{figure}[H]
\centering
\includegraphics[width=\columnwidth]{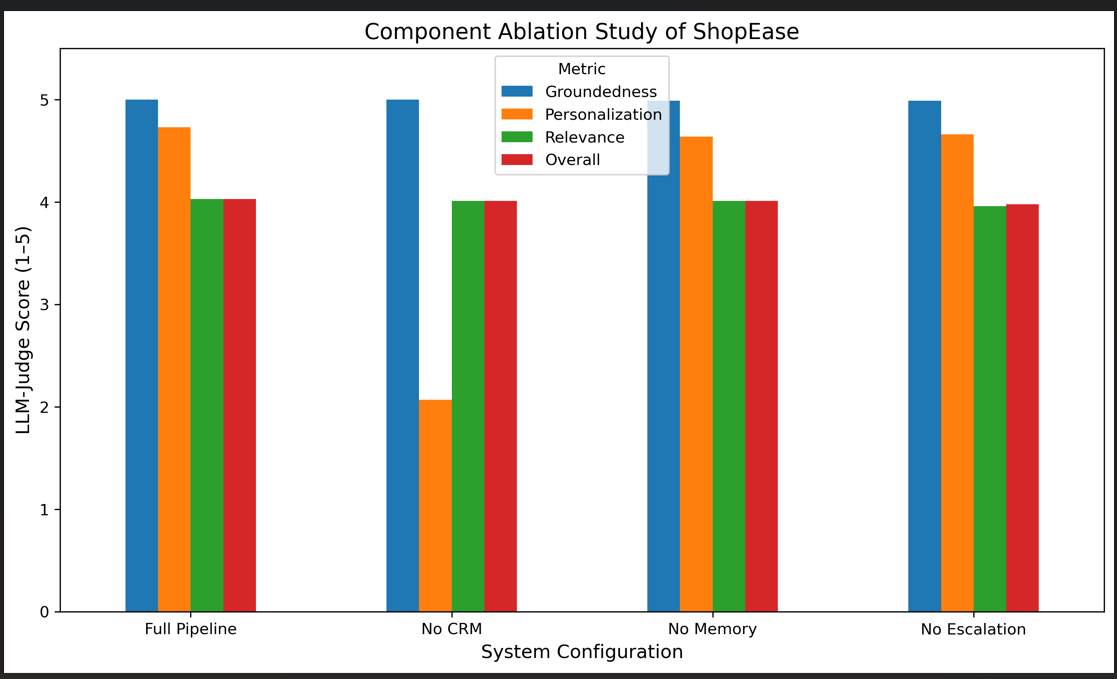}
\caption{Component ablation comparison for the ShopEase workflow.}
\label{fig:component_ablation}
\end{figure}

The component ablation experiment studies the effect of removing selected components from the full ShopEase workflow. Figure~\ref{fig:component_ablation} presents the comparison.
Figure~\ref{fig:component_ablation} compares the full ShopEase pipeline with configurations in which CRM, Memory, or Escalation is removed. The evaluation uses groundedness, personalization, relevance, and overall scores from the LLM-based evaluation.
The Full Pipeline achieves high groundedness and personalization scores while maintaining an overall score of about 4 out of 5. Removing CRM causes the largest drop in personalization, showing that customer information is important for generating personalized responses. Removing Memory also reduces personalization, although the effect is smaller than removing CRM.
The removal of Escalation has a smaller effect on the evaluated response scores. Its main role is to provide a human-review path for cases that require human support rather than directly improving the response scores measured in this experiment.
Overall, the ablation results show that CRM and Memory contribute mainly to the use of customer-specific context, while Escalation provides a separate human-support function within the workflow.
\section{Conclusion and Future Work}
\subsection{Conclusion}
This paper presented ShopEase, a Generative AI-based multi-agent framework for enterprise customer support. The system combines intent detection, CRM information, conversation memory, hybrid policy retrieval, human escalation, and response generation in a single workflow. The system was evaluated on 2632 held-out customer queries using six retrieval configurations. FAISS-only achieved the highest accuracy of 85.37\%, followed closely by Weighted RRF with 85.07\%. BM25-only achieved 55.74\% accuracy. The results show that dense retrieval performed better than sparse retrieval for the evaluation dataset. The experiments also showed that cross-encoder reranking increased retrieval latency without improving accuracy. The error analysis identified Unknown queries and confusion between Refund and Return as the main sources of errors. These results show that retrieval quality has a direct effect on the policy classification performance of the system.
Overall, ShopEase provides a single workflow that combines policy retrieval with customer context, conversation memory, and human escalation. The experimental results show that the system can support enterprise policy-based customer queries while keeping the different support functions within one coordinated framework.

\subsection{Future Work}
Several improvements can be explored in future work. First, the handling of Unknown queries can be improved by adding better out-of-scope detection and stronger rejection criteria for queries that do not match the available policies.
Second, the policy knowledge base can be expanded with more enterprise policies and a larger range of customer queries. This can help evaluate the system on more diverse support scenarios.
Third, retrieval can be further improved by studying better query processing, document chunking, and reranking methods. The effect of these changes can be evaluated using the same held-out evaluation framework.
Finally, the human-in-the-loop component can be extended to support more detailed escalation workflows and feedback collection. Such feedback can be used to improve both retrieval and response generation over time.
\section{Limitations}
The current evaluation is based on a fixed dataset of 2,632 customer queries and a limited set of enterprise policy categories. Therefore, the results may not represent all types of real-world customer-support queries. The system also depends on the quality and coverage of the policy knowledge base. Queries with unclear or out-of-scope information, especially those in the Unknown category, remain challenging. In addition, cross-encoder reranking increases latency without improving accuracy in the current experiments.The current system is evaluated in a controlled experimental setting and does not include long-term deployment data from real customers. The response quality also depends on the retrieved policy information and the local language model. Further evaluation with larger and more diverse datasets would be required to assess the system under broader real-world conditions.
\section*{Declaration of Generative AI and AI-assisted technologies in the writing process}
During the preparation of this work, the authors used Claude (Anthropic) in order to assist with LaTeX formatting, manuscript template preparation, and language editing of the abstract and section text. After using this tool, the authors reviewed and edited the content as needed and take full responsibility for the content of the publication.
\nocite{*}
\bibliographystyle{model5-names}
\biboptions{authoryear}
\bibliography{references}

\end{document}